\documentclass[11pt]{article}

\usepackage{acl}

\usepackage{times}
\usepackage{latexsym}

\usepackage[T1]{fontenc}
\usepackage[utf8]{inputenc}
\usepackage{microtype}
\usepackage{inconsolata}
\usepackage{graphicx}

\usepackage{array}
\usepackage{booktabs}
\usepackage{colortbl}
\usepackage{caption}
\usepackage{amsmath}
\usepackage{amssymb}
\usepackage{xspace}
\usepackage{titletoc}

\usepackage[most]{tcolorbox}
\definecolor{LightGray}{gray}{0.975}
\definecolor{finding1border}{HTML}{1e3a5f}\definecolor{finding1shadow}{HTML}{8ca6cf}
\definecolor{finding2border}{HTML}{5c1a2a}\definecolor{finding2shadow}{HTML}{c87fa3}
\definecolor{finding3border}{HTML}{3d1a5c}\definecolor{finding3shadow}{HTML}{a591bb}
\definecolor{finding4border}{HTML}{1a4d3d}\definecolor{finding4shadow}{HTML}{8cc4b0}
\newtcolorbox{findingbox}[1]{%
  before skip=10pt, after skip=6pt,
  enhanced jigsaw,
  colback=LightGray,
  colframe=finding#1border,
  drop fuzzy shadow southeast={fill=finding#1shadow},
  boxrule=0.9pt, boxsep=0.1pt,
  left=10pt, right=10pt, top=10pt, bottom=10pt}

\newcommand{\bestcell}[1]{\textbf{#1}}

\definecolor{bandretention}{HTML}{DCFCE7}
\definecolor{bandpublished}{HTML}{DBEAFE}

\definecolor{heatred}{HTML}{D1495B}
\definecolor{heatgreen}{HTML}{2E8B57}

\newcommand{\method}{Unified Agent\xspace}
\newcommand{\bench}{\textsc{UA-Bench}\xspace}

\title{Unified Agent: Managing Interactions across Devices}

\author{
Xinshuang Liu \quad Runfa Blark Li \quad Shaoxiu Wei \quad Xin Lin \quad Truong Nguyen \\
University of California, San Diego \\
San Diego, CA, USA \\
\texttt{\{xil235, rul002, shwei, xil321, tqn001\}@ucsd.edu}
}

\begin{document}
\maketitle

\begin{abstract}
As capabilities rapidly increase, AI agents can move from running inside one app to acting across a user's devices over time. Yet existing agent systems still fall short in this scenario. This is because observations are scattered across devices and moments, but mainstream systems are not designed around this fact: a single agent that treats devices as tools lacks effective state management for all devices across time, and multi-agent systems coordinate across agents but do not maintain the compact carried state a cross-device, cross-time request needs. We argue that the agent should maintain an effectively designed state that organizes engagement evidence, stated facts, and the standing request in a compact, action-ready form for deciding its action given the current observation. To compare state designs, we construct a benchmark of user-agent interaction across devices and time. We instantiate this principle in \method{}, a stateful agent that carries interaction evidence across devices and moments and uses it with the current observation to act. In the default setting, it significantly outperforms our adaptations of four published designs. Across changes in multimodal large language model (MLLM) family, capability, and reasoning effort, it remains ahead of all compared systems, demonstrating that the state-design advantage is robust across MLLM settings. Our code and data will be publicly available on GitHub.
\end{abstract}

\section{Introduction} 
\label{sec:intro}

\begin{figure}[t]
\centering
\includegraphics[width=\linewidth]{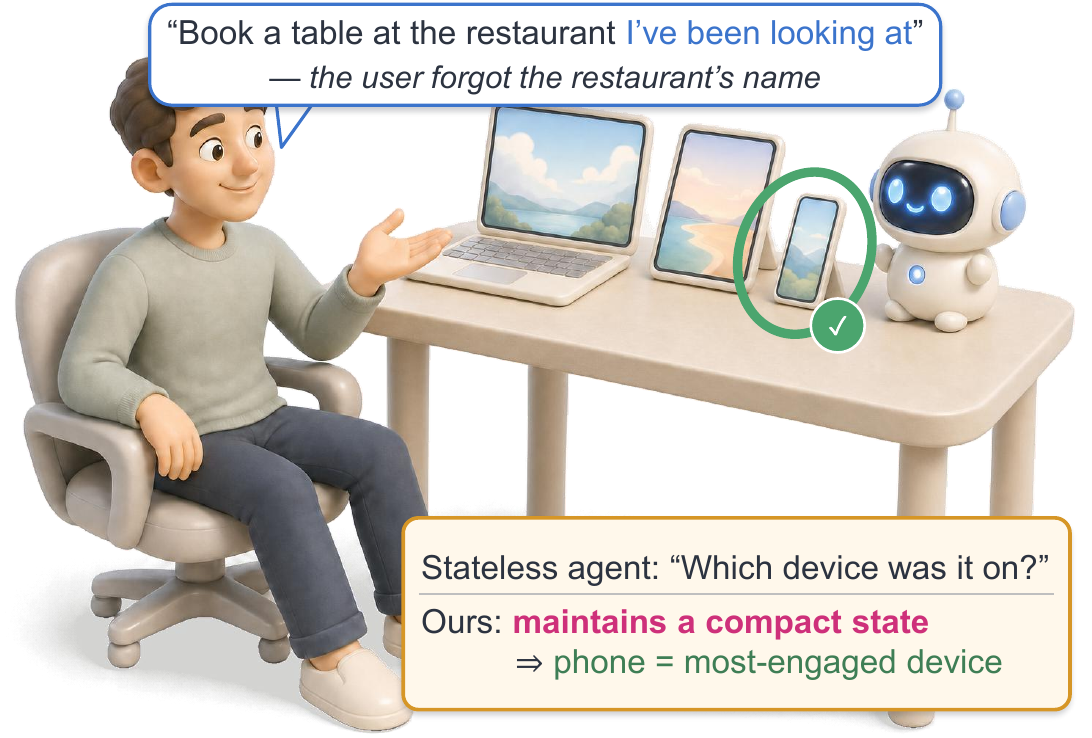}
\caption{\textbf{One agent serving one user across devices and over time.}
After comparing restaurants on several devices, the user later decides to make a reservation but has forgotten the restaurant's name. The user therefore asks to book ``the restaurant I've been looking at'' without specifying a device. A stateless agent may need to ask, ``Which device was it on?'' In contrast, \method{} carries a compact, action-ready state and resolves the reference to the restaurant on the phone, where the user's engagement was strongest. Better state design makes the request answerable.}
\label{fig:teaser}
\end{figure}

As capabilities rapidly increase, AI agents can move from acting inside a single app to managing tasks across a user's devices over time. Figure~\ref{fig:teaser} shows an everyday scenario: a user compares restaurants on several devices, and the agent must later interpret a device-unspecified request. However, today's mainstream agent systems are not designed for this setting. A single-agent system can treat devices as tools, but without an effective cross-time state, it treats each request as fresh; once the deciding cue has passed, the relevant earlier evidence is no longer available. Multi-agent systems can distribute work and coordinate across agents, but without a unified cross-device state, they can still fail when a user's request depends on information spread across devices and moments.

We argue that an agent needs a well-organized state integrating information across devices and moments. In Figure~\ref{fig:teaser}, a stateless agent can only ask, ``Which device was it on?''; an agent whose state has absorbed core interaction evidence across time and devices instead infers that the user engaged most with the phone and books the restaurant there.

We therefore build \method{}, which maintains a state across time and devices. \method{} updates the state with each observation and presents the evidence relevant to the current decision in a compact summary. Each action is selected using both the state and the latest observation. The state records \emph{engagement evidence}, \emph{stated facts}, and the \emph{standing request}. It does not need to store a transcript of everything seen; in the example, retaining that the phone received the strongest engagement is what lets \method{} resolve ``the restaurant I've been looking at'' to the restaurant on the phone. The action-facing summary should hold what a later action may need once the original cue is gone.

To compare state designs, we construct a benchmark of user-agent interaction across time and devices. Our benchmark presents one user interacting with one agent across multiple devices in a 3D scene. Interactions share a designed cross-device sequence across varied scenes, layouts, device placements, topics, and actions, so the correct answers are known by construction and the rendered visual semantics provide ground truth for construction checks. Real-photo results ground the findings in practice. On this benchmark, \method{} significantly outperforms our adaptations of four published designs in the default setting. Across changes in MLLM family, capability, and reasoning effort, it remains ahead of all compared systems, showing that the state-design advantage is robust across MLLM settings. Since \method{} adds no extra mechanism beyond the carried state, the gain traces to how the state is represented and used, not to added system complexity.

Our contributions are summarized as follows:
\begin{itemize}
\item We formulate the problem of user-agent interaction across time and devices.
\item We introduce a unified agent that manages a compact state of engagement evidence, stated facts, and the standing request, and acts based on that state and the latest observation.
\item We construct a benchmark to study user-agent interaction across devices and time.
\item We find that \method{} significantly outperforms our adaptations of four published designs in the default setting and leads all compared systems across MLLM families, capabilities, and reasoning efforts.
\end{itemize}

\begin{figure*}[t]
\centering
\includegraphics[width=0.8\linewidth]{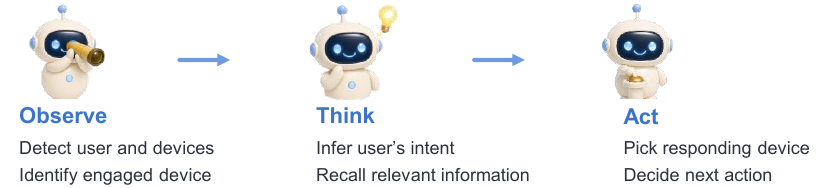}
\caption{\textbf{Six tasks organized as observe--think--act.} The agent perceives the current interaction and identifies the engaged device, reasons with information carried across moments, and determines how to act across devices.}
\label{fig:tasks}
\end{figure*}

\section{Related Work}
\label{sec:related}

\paragraph{Reading the present.}
An agent serving one user across multiple devices often receives only a local observation from a single device camera and the current utterance, while the prior interaction history is not re-supplied. The immediate challenge is therefore attribution: which device is the user engaging with, and what evidence remains once the cue disappears? Prior work infers addressees and engagement from cues such as utterances, gaze, and pointing~\cite{inoue2025addressee,DBLP:conf/mm/YuC0JZD00H25,li2025onlinemmsi,lee2024gazepointar,DBLP:conf/uist/LeeXNQ0CKC0MKD25}; cross-view grounding binds references to perspective~\cite{DBLP:conf/emnlp/TangMS24}. Once the actor or target is identified, device-local and embodied agents can act through screen policies, tool use, or robot affordances~\cite{qin2025uitars,agashe2025agents2,Han2026VLAAGUIKW,acikgoz2025coalm,ichter2023saycan,DBLP:conf/corl/HuangXXCLFZTMCS22,DBLP:conf/icra/LiangHXXHIFZ23,zitkovich2023rt2,react2023}. Situated agents further track intent as a user moves, clarify referents, negotiate roles, or infer household intent over time~\cite{sif2024,DBLP:journals/corr/abs-2508-05535,DBLP:journals/corr/abs-2602-08999,DBLP:journals/corr/abs-2510-23495,DBLP:conf/uist/ArakawaPPLG25}. These works primarily address present-cue interpretation or action once the actor or target has been identified. We study the complementary setting, where engagement evidence must be bound to a physical device and preserved for a later device-unspecified request. \method{} addresses this setting by carrying device-bound engagement evidence beyond the moment in which it is visible.

\paragraph{Carrying state across moments.}
A second line of work examines what should persist across turns. Situated dialogue systems track belief or common ground over shared scenes~\cite{DBLP:conf/naacl/VanderHoevenBKY25,kottur2021simmc2,olvit2024,udagawa2019onecommon,DBLP:journals/corr/abs-2603-12701}, with text-only slot--value tracking providing an earlier form~\cite{budzianowski2018multiwoz}. Agent memory systems store and retrieve extracted facts, evolving notes, temporal graphs, or user profiles~\cite{chhikara2025mem0,xu2025mem,memgpt2023,DBLP:journals/corr/abs-2501-13956,DBLP:journals/corr/abs-2602-13530,DBLP:conf/uist/ShaikhSRHPYB25,DBLP:journals/corr/abs-2509-11914,DBLP:journals/corr/abs-2505-18279}. Recent work also frames shared state as a missing layer over present snapshots~\cite{Wang2026PSISS} and argues for explicit world models as common ground for robots~\cite{DBLP:journals/corr/abs-2601-01705}. Memory benchmarks evaluate recall and update in long conversations, first-person histories, or changing living worlds~\cite{maharana2024locomo,wu2025longmemeval,findingdory2025,Meng2026ClawMarkAL}. These system-oriented methods provide durable stores, typically within a channel, scene, profile, or task, while the benchmarks test whether agents can retain and use information. \method{} instead addresses the cross-device case, where the current observation is insufficient and later decisions require earlier evidence: it preserves per-device engagement evidence, stated facts, and the standing request in one carried state.

\paragraph{Acting across devices.}
A third line of work examines how an intent is executed across devices or agents. Cross-device handoff maintains continuity by moving an artifact, such as a file, clipboard, or assistant embodiment, in response to a sensed cue~\cite{DBLP:conf/chi/JoshiLLPPSLRBMH24,DBLP:conf/socrob/TejwaniKB21}. Orchestrators decompose a given intent and route it through enumerated capabilities, endpoints, or active actors, from capability-routed task graphs to robot teams and smart homes~\cite{zhang2025ufo3,zhang2025agentorchestra,roboosnext2025,Hasan2026M2HRIAL,rivkin2023sage,Zhan2026HearthNetEM}. Embodied multi-agent systems coordinate preassigned agents executors on a shared task~\cite{chang2024partnr,zhang2024coela,liu2025coherent}. These systems provide multi-device control, but typically resolve the target from the present request, a transferred payload, or a static set of registered endpoints rather than from engagement evidence accumulated over time. \method{} instead first infers the relevant device from carried engagement evidence, then recalls the information associated with that device and routes the request. This yields a compact cross-device state that is action-ready for the user's requests, without learned memory, retrieval, or fine-tuning. These properties make the design compatible with different agent systems while supporting efficient integration.

\section{Problem Formulation}
\label{sec:problem}

We study one agent acting for one user across heterogeneous devices. The agent coordinates these devices through a shared decision interface. It uses language to specify which devices should respond and what they should do, while each selected device carries out the instruction using native controls that its manufacturer is best positioned to design and maintain. This separation allows one agent to serve heterogeneous devices while each device retains responsibility for its own execution. The interaction unfolds as a sequence of moments. At each moment, the agent receives one local observation through one device camera, with no global room view and no replay of earlier moments, paired with the user's words. The hard case is a device-unspecified request whose deciding cue has already passed: the engaged screen may be dark, the pointing gesture may be gone, and the request may say only ``the one I have been using.'' The agent must still decide from the current observation and whatever it has carried forward.

We organize each moment around the six observe--think--act tasks shown in Figure~\ref{fig:tasks}: \emph{Detect user and devices}; \emph{Identify engaged device} (Eng); \emph{Infer user's intent} (Int); \emph{Recall relevant information} (Inf); \emph{Pick responding device} (Rsp); and \emph{Decide next action} (Nxt). The first is the perception task; the remaining five are downstream decisions made from the current perception and whatever information the agent carries forward. Rsp is represented as a device set that may be empty, so the agent is not forced to name a device when none applies. Together, the six tasks define the cross-device interaction problem independently of any particular implementation.

The decisions draw on the current view and earlier evidence: which device the user engaged, which facts were stated for that device, and what standing request the user made. A useful cross-device agent therefore needs carried state that is leaner than raw history and more actionable than a list of previous predictions. \method{} carries this evidence forward, so later requests remain answerable when the original cue is no longer visible.

\begin{figure}[t]
\centering
\includegraphics[width=\linewidth]{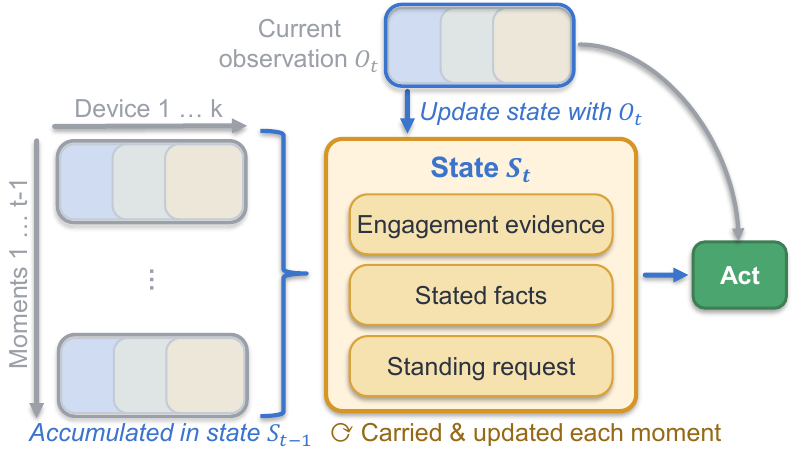}
\caption{\textbf{Overview of \method{}.}
At moment $t$, $S_{t-1}$ carries evidence accumulated across devices over moments $1,\ldots,t-1$. The agent folds the current observation $O_t$ into $S_{t-1}$ to produce $S_t$, which stores engagement evidence, stated facts, and the standing request. It then acts by reading $S_t$ together with $O_t$, and carries $S_t$ forward to the next moment.}
\label{fig:method}
\end{figure}

\section{Method}
\label{sec:method}

An agent acting across a user's devices must decide each action from evidence spread across devices and moments. This evidence may come from an earlier moment or another device and may no longer be available when the agent must act. The agent must therefore carry information forward. Retaining the full history is costly: it grows with the interaction and competes for the model's context window, while a later action typically depends on only a small part of it. The agent needs a compact carried state that preserves the evidence later actions need.

\paragraph{The carried state.}
\method{} meets this need with one carried state $S_t$. It folds each new observation into the state and reads the updated state together with the current observation to act (Figure~\ref{fig:method}). The state keeps \emph{evidence} in three streams, each serving a distinct need of later action. \emph{Engagement evidence} accumulates device-associated interaction cues over time, supporting identification of the relevant device. \emph{Stated facts} preserve user-provided information by topic and, when identifiable, by device, so same-topic facts remain distinguishable across devices. The \emph{standing request} retains the latest requested action. Together, the three streams carry evidence rather than decoded downstream answers. These explicit records are updated with each observation and remain directly available to later decisions.

\paragraph{Updating the state.}
At each moment $t$, the agent folds the current observation $O_t$ into the state:
\begin{equation}
  S_t = U(S_{t-1},\, O_t).
\end{equation}
The update incorporates new evidence while carrying earlier evidence forward. It accumulates engagement evidence, records stated facts with any identifiable device association, and updates the standing request when a new actionable request is recognized. Because folding precedes action, evidence in $O_t$ can affect the current decision.

\paragraph{Acting from the state.}
The agent acts by reading the updated state together with the current observation:
\begin{equation}
  a_t = D(S_t,\, O_t).
\end{equation}
Here, $O_t$ provides evidence from the current moment, while $S_t$ carries evidence accumulated through that moment. Reading both lets evidence from an earlier interaction cue inform action after the cue is no longer present in $O_t$, while new evidence can revise the decision. The agent uses engagement evidence to identify the relevant device and stated facts to recover information when needed. Device capabilities are supplied separately rather than carried in $S_t$ and are omitted from the notation. The decoder obtains the action from the standing request and combines it with engagement evidence and device capabilities to decide which device or devices should respond. \method{} communicates the resulting instruction in language through the shared interface, while each selected device carries it out using its native controls. This separation is consistent with prior work on tool use and robot skills~\cite{schick2023toolformer,ichter2023saycan}.

\section{Benchmark}
\label{sec:benchmark}

Our \bench{} is a reproducible benchmark of interaction across devices and time. \bench{} targets devices in a 3D space. The agent must combine local device-camera views of user--device interaction with the user's words before deciding. Controlled rendering preserves the need for visual and spatial reasoning about situated device interaction. As in visual and long-context benchmarks~\cite{johnson2017clevr,kuratov2024babilong}, it also provides exact semantic ground truth for construction checks and supports systematic variation and reproducibility.

\begin{figure}[t]
\centering
\begin{minipage}{0.485\linewidth}\centering
\includegraphics[width=\linewidth]{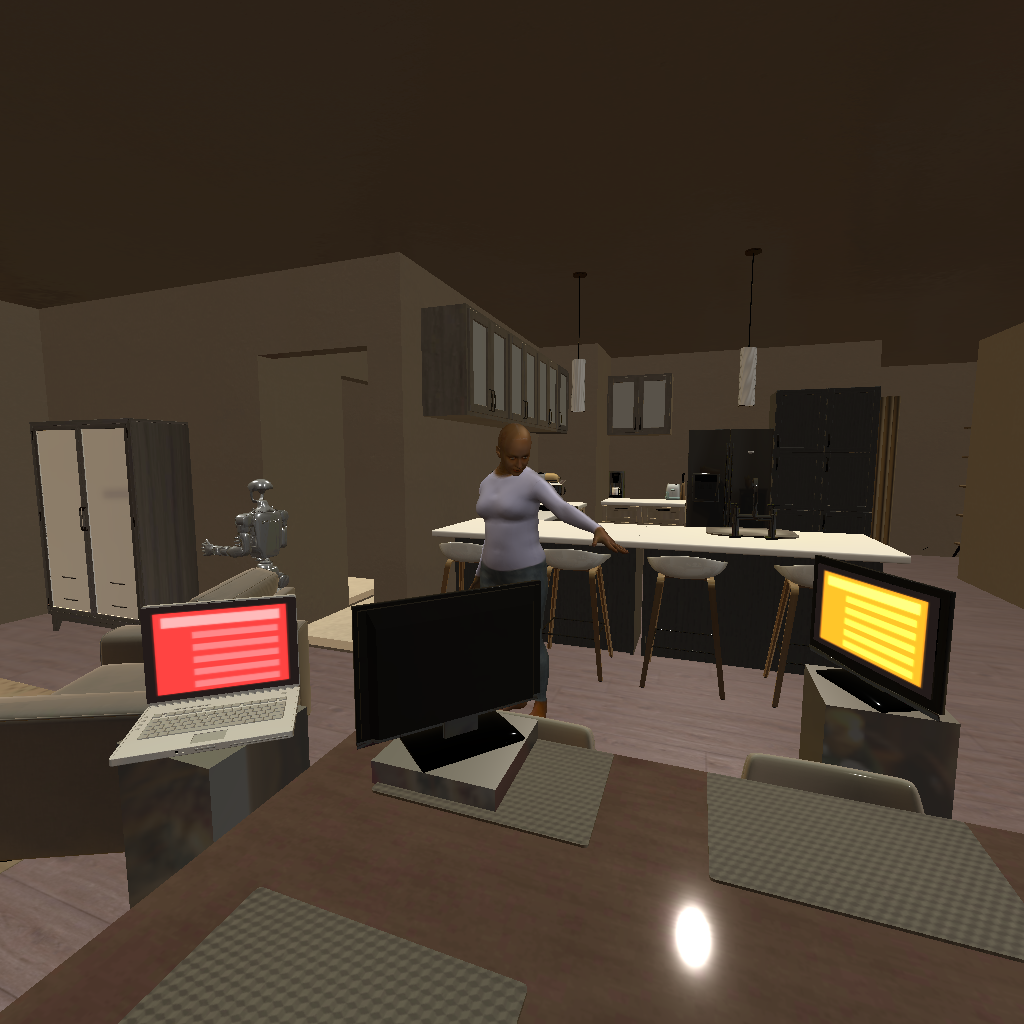}\\[1pt]
{\footnotesize (a) Device-camera view}
\end{minipage}\hfill
\begin{minipage}{0.485\linewidth}\centering
\includegraphics[width=\linewidth]{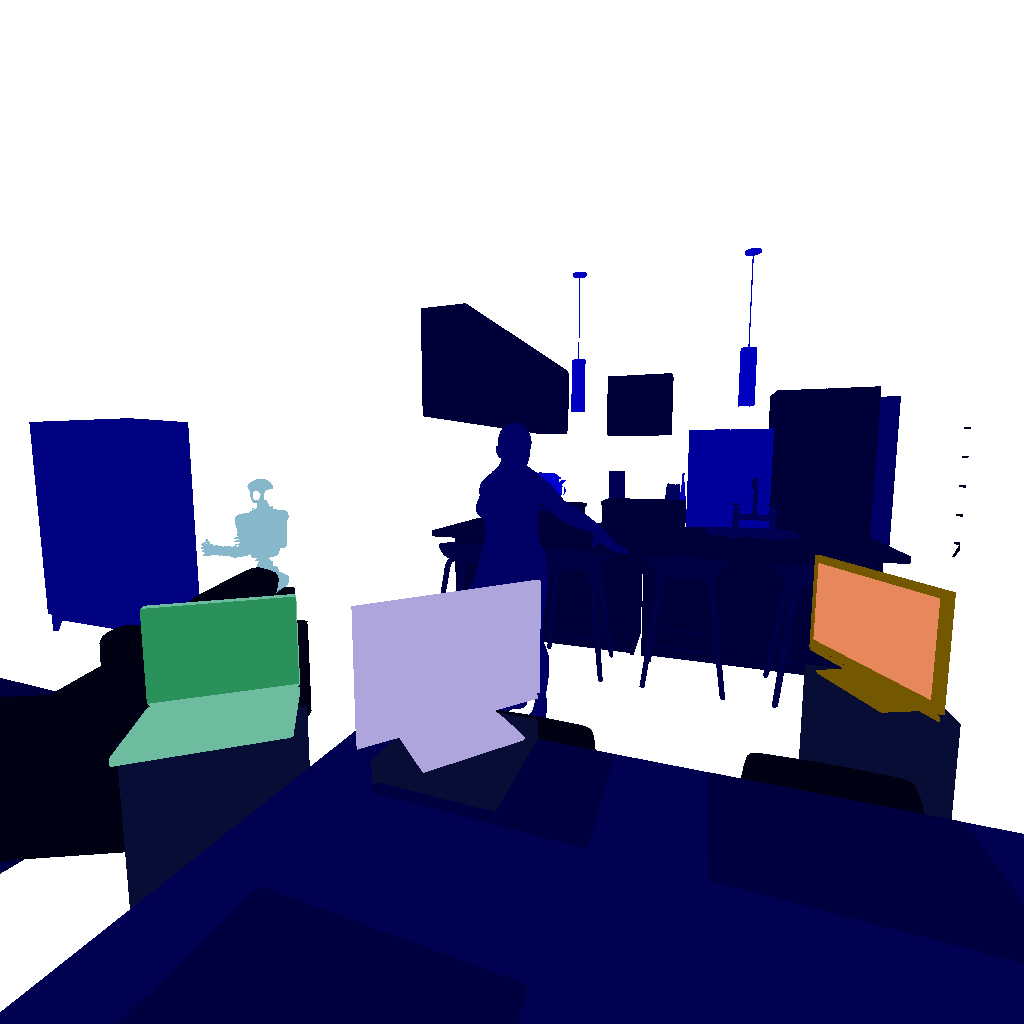}\\[1pt]
{\footnotesize (b) Semantic map}
\end{minipage}
\caption{\textbf{Rendered \bench{} frame.} (a) An agent-visible device-camera view shows user--device interaction in a scene containing screen devices and a fetch robot. (b) The paired semantic ground truth supports construction-time validation of object identity and visibility and is not available to evaluated systems.}
\label{fig:bench-example}
\end{figure}

An episode is a structured cross-device interaction in a rendered 3D home~\cite{DBLP:conf/iclr/PuigUSCYPDCHMVG24,khanna2024hssd}, observed through device cameras and paired at each frame with the user's words (Figure~\ref{fig:bench-example}). The cast includes a rendered user and screen devices such as a computer, a laptop, and a phone; episodes may additionally field a fetch robot, which extends the responding-device choices beyond screens. Each episode follows a designed temporal progression: an initial observation establishes the scene; an earlier interaction establishes user--device engagement and associated context through visual and linguistic cues; and a later device-unspecified request and recall questions test whether that information remains available as the immediate visual evidence changes. \bench{} instantiates the six tasks of Section~\ref{sec:problem}. Device-camera observations provide the visual input for \emph{Detect user and devices}; the five downstream targets are \emph{Identify engaged device} (Eng), \emph{Infer user's intent} (Int), \emph{Recall relevant information} (Inf), \emph{Pick responding device} (Rsp), and \emph{Decide next action} (Nxt). Evaluation targets are determined directly from the controlled episode specification rather than from model-generated annotations, so every target is exact by construction.

Benchmark construction information is strictly separated from evaluated-system inputs. Semantic ground truth and render metadata are used only during benchmark construction to establish exact ground truth for entity visibility and validate object identity and screen state. System inputs are derived solely from rendered device-camera views, the user's words, and a public description of the task and devices, including device identities, appearances, and capabilities; the controlled episode specification and evaluation targets remain unavailable to evaluated systems. Together, these system inputs and evaluation targets provide a common basis for comparing different cross-device agent designs. Within each MLLM setting, all MLLM-based methods are evaluated using a shared per-frame perception output. When the decoder is also shared, the comparison isolates differences in state design while preserving \method{}'s perceive--fold--act pipeline.

\bench{} contains 100 matched pairs (200 samples) spanning varied indoor scenes, layouts, device placements, topics, and actions. Each pair varies the preceding interaction under a common task structure and evaluation protocol, enabling direct comparison of how systems use information carried from earlier moments. Every admitted episode passes automatic checks of construction consistency, user and candidate-device visibility, and matched-pair integrity. Rendered batches also undergo human review for image legibility and rendering quality. Appendix~\ref{app:construction} documents the construction and admission protocol. Section~\ref{sec:realcase} complements the rendered benchmark with photographs.

\begin{table*}[t]
\centering
\small
\caption{\textbf{State-design comparison.} Per-system scores for the five downstream decisions on GPT-5.6-Luna~\cite{openai2026gpt56} (low reasoning effort); bold marks the best overall performance. With its compact, action-ready state, \method{} achieves the highest overall performance among the compared systems without retaining raw history.}
\label{tab:headline}
\setlength{\tabcolsep}{5.5pt}
\begin{tabular}{l *{5}{c} @{\hspace{8pt}}c r}
\toprule
& \multicolumn{5}{c}{Decision score ($\uparrow$)} & & \\
\cmidrule(lr){2-6}
System & Eng & Int & Inf & Rsp & Nxt & \textbf{Overall} & Gap [95\% CI] \\
\midrule
\textbf{\method{} (Ours)} & 0.777 & 0.307 & 0.745 & 0.734 & 0.777 & \bestcell{0.668}\,{\scriptsize[.64,.70]} &  \\
\midrule
\rowcolor{bandretention}\multicolumn{8}{l}{\textit{State controls}} \\
Full context & 0.765 & 0.338 & 0.735 & 0.757 & 0.472 & 0.613\,{\scriptsize[.58,.64]} & +0.055\,{\scriptsize[+0.035, +0.075]} \\
Self-notes & 0.760 & 0.331 & 0.207 & 0.736 & 0.367 & 0.480\,{\scriptsize[.46,.50]} & +0.188\,{\scriptsize[+0.160, +0.214]} \\
Answer cache & 0.172 & 0.321 & 0.000 & 0.668 & 0.057 & 0.244\,{\scriptsize[.23,.26]} & +0.424\,{\scriptsize[+0.389, +0.459]} \\
Text only & 0.035 & 0.369 & 0.000 & 0.647 & 0.000 & 0.210\,{\scriptsize[.20,.22]} & +0.458\,{\scriptsize[+0.426, +0.489]} \\
Observation only & 0.032 & 0.354 & 0.000 & 0.605 & 0.000 & 0.198\,{\scriptsize[.19,.21]} & +0.470\,{\scriptsize[+0.437, +0.502]} \\
\midrule
\rowcolor{bandpublished}\multicolumn{8}{l}{\textit{Published baselines}} \\
Mixture-of-Agents & 0.680 & 0.299 & 0.480 & 0.697 & 0.215 & 0.474\,{\scriptsize[.45,.50]} & +0.194\,{\scriptsize[+0.161, +0.226]} \\
Debate-or-vote & 0.703 & 0.301 & 0.435 & 0.689 & 0.242 & 0.474\,{\scriptsize[.45,.50]} & +0.194\,{\scriptsize[+0.161, +0.225]} \\
Mem0 & 0.082 & 0.304 & 0.395 & 0.624 & 0.035 & 0.288\,{\scriptsize[.27,.31]} & +0.380\,{\scriptsize[+0.338, +0.420]} \\
MM-DST & 0.195 & 0.271 & 0.025 & 0.649 & 0.162 & 0.260\,{\scriptsize[.24,.28]} & +0.408\,{\scriptsize[+0.363, +0.451]} \\
\bottomrule
\end{tabular}

\par\vspace{2pt}\raggedright\footnotesize Appendix~\ref{app:systems} details the setup of each compared system. \textbf{Overall} is the macro-average of the five downstream decisions, giving each required decision equal weight regardless of its number of scored instances; its brackets show the one-sample 95\% bootstrap interval clustered by matched pair (Appendix~\ref{app:stats}). \textbf{Gap} is the difference between \method{}'s Overall score and that of each comparison system. Brackets report the corresponding paired 95\% interval; positive values favor \method{}. All gaps remain significant after Holm adjustment within their respective comparison families ($p<10^{-3}$).
\end{table*}

\section{Experiments}
\label{sec:experiments}

\providecommand{\yes}{\ensuremath{\checkmark}}
\providecommand{\no}{\ensuremath{\times}}
\providecommand{\hf}{\ensuremath{\circ}}

We first compare \method{} with state controls and with memory and multi-agent baselines. We then test how key design elements support later decisions, whether the advantage generalizes across MLLM settings, and whether the same principle holds on real photographs.

\subsection{Experimental setup}

\begin{figure}[t]
\centering
\includegraphics[width=\linewidth]{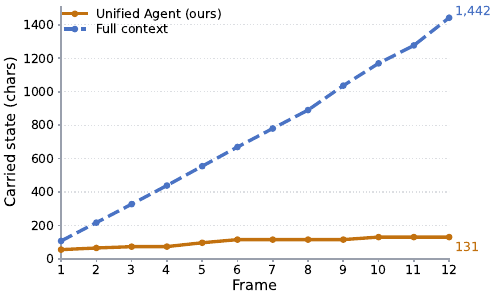}
\caption{\textbf{Carried-state size.} Characters of retained values after each frame, averaged over the 200 episodes; fixed template text is excluded identically for every system. \method{} stays bounded; Full context grows linearly to 11 times the size of \method{}'s state.}
\label{fig:state-size}
\end{figure}

We compare methods under a common evaluation harness. Within each MLLM setting, all MLLM-based comparison systems use fixed zero-shot prompt templates and share the same perception record, model, final output schema, and scorer, while each retains its own mechanism for representing and carrying information. Appendices~\ref{app:eval-scoring} and~\ref{app:systems} detail the evaluation protocol, model and inference settings, output schema, and baseline implementations used in the reported comparisons.
We evaluate four MLLM settings that vary in reasoning effort, model family, and model capability. Unless otherwise noted, GPT-5.6-Luna~\cite{openai2026gpt56} at low reasoning effort is the default setting. Relative to the default, GPT-5.6-Luna at medium reasoning effort varies the reasoning setting; Gemini 3.1 Flash-Lite~\cite{googledeepmind2026gemini31flashlite} at low thinking level varies the model family; and GPT-5.6-Sol~\cite{openai2026gpt56} at low reasoning effort varies model capability. Within each MLLM setting, every MLLM-based method receives a common per-frame perception output. The methods differ in what information they carry forward and how they use it. This protocol holds the perception input constant across methods, enabling a direct comparison. Table~\ref{tab:perception} in Appendix~\ref{app:eval-store} reports the quality of shared perception across settings. \method{} continues to perceive, fold each observation into its state, and act from the updated state.

We report visual-perception diagnostics supporting \emph{Detect user and devices} separately from the five downstream decisions. Holding perception fixed within each MLLM setting, the comparison scores Eng, Int, Inf, Rsp, and Nxt. Engagement, recall, routing, and next action are graded deterministically against by-construction ground truth; intent is scored by a fixed same-meaning equivalence MLLM judge (Appendix~\ref{app:eval-scoring-rules}). Uncertainty is quantified by a bootstrap clustered on matched pairs (Appendix~\ref{app:stats}).

\subsection{State-design comparison}

\begin{findingbox}{1}
{\textbf{Finding 1:} A well-organized carried state provides a performance advantage for cross-device agents: \method{} achieves the highest overall score among all compared systems.}
\end{findingbox}

This subsection examines a central question in state design: how accumulated information should be represented and used for later action. We compare \method{} with the five state controls of Table~\ref{tab:headline} and with four published-design baselines (Appendix~\ref{app:systems-controls}):
\begin{itemize}
\item \textbf{Memory and dialogue-state methods:} Mem0~\cite{chhikara2025mem0} and MM-DST (multimodal dialogue state tracking)~\cite{vdtn2022,olvit2024}.
\item \textbf{Multi-agent methods:} Mixture-of-Agents~\cite{wang2025moa} and Debate-or-vote~\cite{choi2025debate}.
\end{itemize}

Table~\ref{tab:headline} shows that \method{} leads all compared systems in overall performance, including the strongest control, Full context. A paired bootstrap clustered on matched pairs places a 95\% confidence interval on every reported gap, and each gap remains significant after Holm correction within its comparison family (all adjusted $p<10^{-3}$; Appendix~\ref{app:stats}). Full context, which retains all prior perception records and utterances from which \method{} derives its state, scores slightly higher on intent and routing, making it a strong control. Figure~\ref{fig:state-size} reports the size of the carried state across the interaction for both designs. Full context grows linearly because it retains every prior frame; \method{}'s state scales with the observed devices and stated facts rather than with the interaction length, and therefore remains bounded. \method{} achieves the highest overall performance, with its largest advantage on next action.

Across individual decisions, \method{} leads every published design on engagement, recall, routing, and next action. Mem0 and MM-DST manage cross-frame information through retrieved fact memory and updated dialogue state, while Mixture-of-Agents and Debate-or-vote coordinate device agents through proposal synthesis and peer-feedback voting. Observation only decodes from the current shared perception record; its contrast with \method{} demonstrates the value of carrying engagement evidence across frames.

\begin{table}[!t]
\centering
\small
\caption{\textbf{Ablations of key design elements.} Each row reports the signed difference between a variant's per-decision score and the corresponding \method{} score in Table~\ref{tab:headline}; negative values favor \method{}, and positive values favor the variant. Shading indicates the absolute magnitude. All variants use the same fixed per-frame record and differ from \method{} only in the specified carried-state, visual-evidence, or routing component. Appendix~\ref{app:systems-ablations} defines each variant.}
\label{tab:necessity}
\setlength{\tabcolsep}{2pt}
\begin{tabular}{@{}l *{5}{c}@{}}
\toprule
 & Eng & Int & Inf & Rsp & Nxt \\
\midrule
Count-free evidence & \cellcolor{heatred!55!white}-0.74 & \cellcolor{heatgreen!3!white}0.04 & \cellcolor{heatred!46!white}-0.61 & \cellcolor{heatred!11!white}-0.15 & \cellcolor{heatred!55!white}-0.74 \\
No pointing channel & \cellcolor{heatred!55!white}-0.76 & \cellcolor{heatgreen!4!white}0.05 & \cellcolor{heatred!17!white}-0.23 & \cellcolor{heatred!3!white}-0.03 & \cellcolor{heatred!55!white}-0.76 \\
Reduced-ability routing & 0.00 & \cellcolor{heatgreen!1!white}0.01 & 0.00 & \cellcolor{heatred!14!white}-0.18 & 0.00 \\
Frame-local state & \cellcolor{heatred!55!white}-0.77 & \cellcolor{heatgreen!3!white}0.04 & \cellcolor{heatred!55!white}-0.74 & \cellcolor{heatred!13!white}-0.18 & \cellcolor{heatred!55!white}-0.77 \\
\bottomrule
\end{tabular}

\end{table}

\begin{table}[t]
\centering
\small
\caption{\textbf{Performance across MLLM settings.} Overall performance across the five downstream decisions for \method{} and the five state controls; Gap values are computed from unrounded scores; within each column the MLLM-based systems share the same MLLM and inference setting. Bold marks the best value in each column, and Gap is \method{}'s overall performance minus that of the strongest control. The state advantage persists across MLLM family, model capability, and reasoning effort.}
\label{tab:generalization}
\setlength{\tabcolsep}{3pt}
\begin{tabular}{@{}l c >{\columncolor{gray!12}}c c c@{}}
\toprule
 & \multicolumn{1}{c}{\shortstack{Gemini 3.1\\ Flash-Lite}} & \multicolumn{2}{c}{\shortstack{GPT-5.6-\\ Luna}} & \multicolumn{1}{c}{\shortstack{GPT-5.6-\\ Sol}} \\
\cmidrule(lr){2-2}\cmidrule(lr){3-4}\cmidrule(lr){5-5}
System & low & low & medium & low \\
\midrule
\textbf{\method{}} & \bestcell{0.476} & \bestcell{0.668} & \bestcell{0.670} & \bestcell{0.715} \\
\midrule
Full context & 0.445 & 0.613 & 0.611 & 0.683 \\
Self-notes & 0.402 & 0.480 & 0.485 & 0.590 \\
Answer cache & 0.382 & 0.244 & 0.232 & 0.294 \\
Text only & 0.201 & 0.210 & 0.209 & 0.219 \\
Observation only & 0.197 & 0.198 & 0.198 & 0.185 \\
\midrule
Gap & +0.031 & +0.055 & +0.058 & +0.033 \\
\bottomrule
\end{tabular}

\end{table}

\subsection{Ablations of key design elements}

\begin{findingbox}{2}
{\textbf{Finding 2:} Different tasks need different information: ablations of \method{} show that engagement evidence supports identifying the engaged device, prior state supports recall, and device abilities support routing.}
\end{findingbox}

Table~\ref{tab:necessity} evaluates four controlled variants. Count-free evidence removes multiplicity from activity and pointing evidence; No pointing channel removes pointing from carried and current-frame inputs; Reduced-ability routing removes only the per-device ability clauses; and Frame-local state resets prior state before folding the unchanged current frame. The targeted contrasts show losses on Eng without counts or pointing, on Rsp without per-device abilities, and on Inf without prior state. All four targeted gaps are significant after Holm correction (all adjusted $p<10^{-3}$; Appendix~\ref{app:stats}).

Within each MLLM setting, the compared state designs use the same perception record, whose quality is evaluated separately in Table~\ref{tab:perception}. Because the controlled variants share this record, their contrasts isolate the contribution of state and evidence pathways after perception. The results distinguish complementary roles for carried information: carried engagement cues identify the engaged device, retained facts support recall, and device constraints guide routing. Together, these mechanisms connect earlier interaction evidence to later decisions.

\begin{figure*}[t]
\centering
\includegraphics[width=\textwidth]{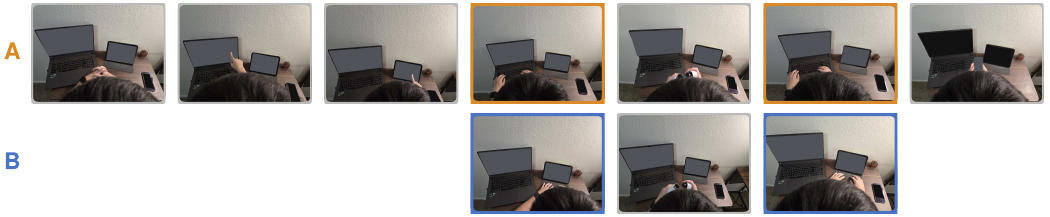}
\caption{\textbf{Real-photo matched pair.} The user engages with the laptop in episode A and the tablet in episode B; the shared request frame is identical across the pair while the earlier engagement is swapped.}
\label{fig:app-realcase}
\end{figure*}

\subsection{Generalization across MLLM settings}

\begin{findingbox}{3}
{\textbf{Finding 3:} The benefit of a well-organized state generalizes across MLLM settings: \method{} outperforms all compared systems across MLLM families, capabilities, and reasoning efforts.}
\end{findingbox}

This subsection tests whether the state-design advantage generalizes across MLLM settings. Table~\ref{tab:generalization} evaluates four settings spanning MLLM families, model capabilities, and reasoning efforts and reports overall performance for \method{} and the five state controls.

In every evaluated setting, \method{} achieves the highest overall performance among the compared systems and leads Full context, the strongest control, with the ordering preserved across MLLM families, model capabilities, and reasoning efforts.

\subsection{Carried state on real photographs}
\label{sec:realcase}

\begin{findingbox}{4}
{\textbf{Finding 4:} Carried state supports cross-device decisions from real photographs: \method{} identifies the engaged device and recalls the relevant information.}
\end{findingbox}

\begin{table}[t]
\centering
\small
\caption{\textbf{Real-photograph case study.} Per-decision outcomes under GPT-5.6-Luna with low reasoning effort. \yes{} denotes a correct result in both episodes; \hf{} denotes either a correct result in one episode or a partially correct decision; and \no{} denotes an incorrect result. Appendix~\ref{app:realcase} provides episode-level details.}
\label{tab:realcase}
\setlength{\tabcolsep}{3pt}
\begin{tabular}{l ccccc}
\toprule
 & Eng & Int & Inf & Rsp & Nxt \\
\midrule
\textbf{\method{} (Ours)} & \yes & \hf & \yes & \yes & \yes \\
\midrule
Full context & \yes & \hf & \no & \yes & \yes \\
Self-notes & \yes & \hf & \no & \yes & \yes \\
Answer cache & \yes & \hf & \hf & \yes & \yes \\
Text only & \no & \hf & \no & \yes & \yes \\
Observation only & \no & \hf & \no & \yes & \yes \\
\bottomrule
\end{tabular}

\end{table}

We test the same state principle with real photographs. In a home-desk scenario, a person uses a laptop, tablet, and phone to book a family birthday dinner (Figure~\ref{fig:app-realcase}). They compare restaurants with different stated times, work mainly on one device, then later pick up their phone and ask, with both work screens dark and without naming a device, to text everyone the time for ``the one I was just working on.''

In the case study, \method{} runs its full perceive--fold--act pipeline directly on the photographs, with hand-on-device contact as the engagement cue. Throughout the real-photo evaluation, it identifies the device with the strongest accumulated engagement from its carried state and recalls the time stated for that device rather than the decoy time (Table~\ref{tab:realcase}; Appendix~\ref{app:realcase}).

\section{Conclusion}
\label{sec:conclusion}

Cross-device agents can fail when a request depends on evidence scattered across devices and moments but no longer visible at the time of action. We argued that such agents need a compact carried state, and we instantiated this design in \method{}, whose state stays bounded as the interaction grows. In the experiments, \method{} achieves the highest overall performance among the compared systems, ahead of Full context and the memory and multi-agent baselines. Against the four published baselines, Overall gains span 0.194--0.408, with all paired-bootstrap 95\% intervals above zero. The controlled ablations further show that evidence counts, pointing, prior state, and device abilities each support their targeted downstream decisions. The advantage persists across MLLM settings. The findings highlight effective state design as an important complement to model strength for cross-device agents.

\section{Limitations}

Privacy in any stateful design turns on how much user context is carried across devices and time; designs that keep the raw interaction record retain the most. That record also grows with the interaction, crowding the model's context, whereas \method{} uses its carried state to present a compact, action-facing summary of engagement evidence, stated facts, and the standing request for later decisions. Because this state is explicit rather than latent, retained information remains directly inspectable and selectively revisable, allowing the record to be minimized without losing the context needed to answer the user's requests.

\bench{} provides exact, traceable evaluation targets for cross-device interaction. Its structured targets derive directly from each episode's design and device metadata, without model annotation. Render metadata and semantic maps provide construction-time validation and remain unavailable to evaluated systems. The resulting structured ground truth is exact rather than estimated and is verified by deterministic recovery checks, while batch-level human review screens the rendered images for legibility and rendering quality. A matched-pair real-photo case study complements the rendered benchmark.

\section{Ethical Considerations}

Agents that manage interactions across a user's devices can lower the everyday effort of coordination, keeping a request answerable when the screens are dark and no device is named, and sparing the user the work of reconstructing which device held what. Keeping the selected device, recalled information, and proposed action aligned with the user's intent is central to dependable cross-device coordination. \method{} keeps these elements explicit, so they can be reviewed together before execution. To preserve privacy across devices, \method{} retains a compact, explicit record for later action instead of the complete interaction history: data minimization by design. Engagement signals relevance, not identity or authorization, particularly on shared devices. Uncertain or stale evidence should therefore prompt confirmation before information or instructions are routed. Our code and data will be released publicly, supporting reproducibility and further development of carried-state designs.

\bibliography{main}

\clearpage

\appendix

\startcontents[appendix]
\printcontents[appendix]{l}{1}{\section*{Appendix Contents}}

\section{Carried State Specification}
\label{app:state-task}

\subsection{State representation}
\label{app:state-streams}

For an episode with device set $\mathcal{D}$, let $\mathcal{D}_{\mathrm{eng}}\subseteq\mathcal{D}$ be the screen devices eligible to be labeled as the engaged device. After frame $t$, \method{} carries
\begin{equation}
  S_t = (C_t, P_t, K_t, r_t),
\end{equation}
where $C_t$ and $P_t$ encode engagement evidence, $K_t$ stores stated facts, and $r_t$ stores the standing request. In $C_t$, $m_t$ counts nonempty engagement observations, activity in the rendered benchmark and attended-device contact in the real-photo case, and $\mathrm{lit}_t(d)$ counts those containing $d$, for each $d\in\mathcal{D}_{\mathrm{eng}}$. $P_t$ retains attended-device cues: pointing in the rendered benchmark and hand-on-device contact in the real-photo case. The decision interface summarizes these cues as per-device tallies. $K_t$ stores the first stated fact for each device--topic pair when the device is known and for each topic otherwise. $r_t$ stores the latest requested action and, in the real-photo case, the queried topic.

\subsection{State update and decision interface}
\label{app:state-update}

The state evolves by folding in each observation,
\begin{equation}
  S_t = U(S_{t-1},\, O_t).
\end{equation}
For each $O_t$, activity and attended-device cues update engagement evidence; a stated fact at a nonempty activity frame is added only when no fact is already recorded for that device and topic; and an eligible command replaces the standing request. Other components persist. The agent folds $O_t$ into $S_{t-1}$ before predicting from $S_t$ and the current observation.

Eng uses accumulated activity and attended-device cues; Int interprets the current utterance; Inf retrieves the fact for the queried topic, resolving same-topic facts by device; Rsp combines the requested action, engagement evidence, and card-declared device abilities; and Nxt combines the standing request with engagement evidence. The decoder receives the per-device activity and cue tallies, one record for each stored fact, the standing request, and the current observation.

\section{Benchmark Design and Construction}
\label{app:construction}

\bench{} consists of controlled matched pairs in which later decisions depend on evidence from earlier interaction frames. This section presents the benchmark scope, matched-pair design, and episode construction and admission.

\subsection{Benchmark scope}
\label{app:construction-scope}

Each episode contains a temporally ordered interaction among one user, candidate screen devices, and, in episodes that include it, a fetch robot. Table~\ref{tab:construction-summary} summarizes the benchmark scale, frame structure, device roles, and recall vocabulary.

\begin{table*}[t]
\centering
\small
\caption{\textbf{\bench{} structure and coverage.} Dataset scale, episode timeline, device roles, and recall vocabulary.}
\label{tab:construction-summary}
\begin{tabular}{@{}ll@{}}
\toprule
Component & Specification \\
\midrule
Matched pairs & 100 \\
Episodes & 200 (two per matched pair) \\
Episode length & 12 frames \\
Shared frames & $f_1, f_{10}, f_{11}, f_{12}$ \\
Role-swapped engagement frames & $f_2$--$f_9$ \\
Candidate screen-device categories & laptop, computer \\
Service device & fetch robot (episode-dependent) \\
Recall vocabulary & 8 topics with 2 values each \\
\bottomrule
\end{tabular}
\end{table*}

\subsection{Matched-pair design}
\label{app:construction-pair}

\begin{figure*}[t]
\centering
\includegraphics[width=\linewidth]{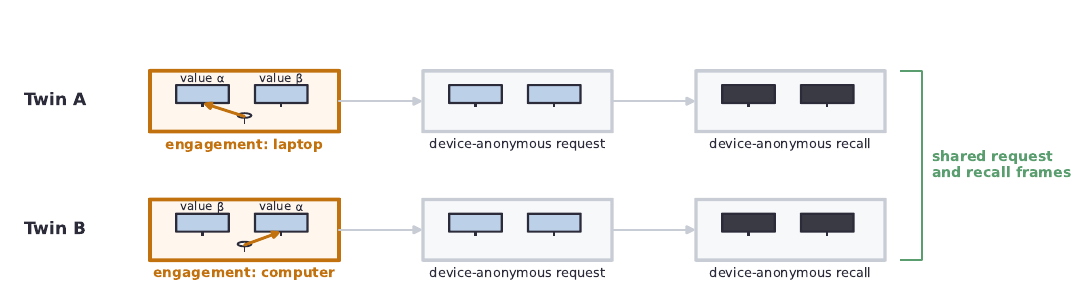}
\caption{\textbf{Matched-pair design.} The paired episodes exchange earlier engagement roles and associated values while sharing the later request and recall frames. The paired construction makes earlier interaction evidence the distinguishing context for the later decisions.}
\label{fig:mechanism}
\end{figure*}

Each matched pair contains two episodes with identical opening, request, and recall frames (Table~\ref{tab:construction-summary}). The intervening engagement frames exchange the engaged and decoy roles and their associated values, yielding different Eng, Inf, Rsp, and Nxt targets while preserving Int across the role swap. Figure~\ref{fig:mechanism} summarizes this construction.

\subsection{Episode construction and admission}
\label{app:construction-record}

Each episode is scripted in a rendered 3D household scene containing one user, screen devices, additional screen distractors, and, in episodes that field it, a MuJoCo Menagerie humanoid robot (Unitree G1)~\cite{menagerie2022github} in the fetch-robot role.

\paragraph{Agent inputs.} The agent receives temporally ordered device-camera images and utterances, a device roster, and a device card. The roster specifies device identity, category, appearance, and any screen-glow color; the card specifies the device set, each device's fixed abilities, the observation rule, and the closed answer vocabularies.

\paragraph{Construction records.} Each episode's construction record comprises the scene structure, engagement schedules, role assignments, render metadata, semantic maps, pair identities, and target labels, supporting benchmark generation and validation. Agent-facing episodes contain only the inputs defined above.

\paragraph{Ground truth and admission.} Eng, Inf, Rsp, and Nxt are deterministic functions of the construction script, device metadata, and controlled role assignment; Int uses an authored reference validated before admission. Admission requires recoverable targets, designated-frame constraints on vocabulary and actions, leakage and device-naming checks, image-level checks of pointing-gesture and candidate-device visibility, and the matched-pair invariants of the preceding subsection. Rendered batches additionally undergo manual screening for legibility and rendering quality.

The construction enforces the following invariants for every admitted pair.

\begin{enumerate}
\item \textbf{Shared-frame identity.} The shared frames are rendered once and reused, making them identical across the pair.

\item \textbf{Role swap.} The engaged and decoy roles are exchanged over the engagement frames, so the engaged-device label differs across the pair.

\item \textbf{Value collision.} The same topic receives distinct values on two devices. Statement order is reversed within each pair and balanced across pairs.

\item \textbf{Anonymous wording.} Utterances use device-unspecified wording, and recall questions omit the queried value.

\item \textbf{Ability-constrained routing.} Responding-device targets follow the card-declared abilities: a device supports a request exactly when its card covers the requested action, so targets range over device sets, including a single device and the empty set, while engagement targets are defined over the candidate screen devices.
\end{enumerate}

\begin{table*}[t]
\centering
\small
\begin{minipage}[t]{0.48\textwidth}
\vspace{0pt}
\centering
\caption{\textbf{Downstream output specification.} Output space and scoring rule for each evaluated decision.}
\label{tab:contract}
{\footnotesize
\begin{tabular}{l@{\hspace{6pt}}l@{\hspace{6pt}}l}
\toprule
Decision & Output space & Scoring rule \\
\midrule
Eng & device identifier or none & exact match \\
Int & free-form text & semantic match \\
Inf & value, unknown, or none & case-insensitive match \\
Rsp & set of device identifiers & set-$F_1$ \\
Nxt & action--device pair & exact match \\
\bottomrule
\end{tabular}}
\end{minipage}
\hfill
\begin{minipage}[t]{0.48\textwidth}
\vspace{0pt}
\centering
\caption{\textbf{Evaluated MLLM settings.} GPT-5.6-Luna at low reasoning effort is the primary setting.}
\label{tab:mllm-settings}
\setlength{\tabcolsep}{4pt}
\begin{tabular}{lll}
\toprule
MLLM & Effort & Purpose \\
\midrule
Gemini 3.1 Flash-Lite & low & cross-family \\
GPT-5.6-Luna & low & primary setting \\
GPT-5.6-Luna & medium & cross-effort \\
GPT-5.6-Sol & low & cross-capability \\
\bottomrule
\end{tabular}
\end{minipage}
\end{table*}

\section{Evaluation Protocol and Scoring}
\label{app:eval-scoring}

Within each MLLM setting, the evaluation fixes the shared perception record, output specification, and scorer. This section specifies the output specification, MLLM settings, shared records, and scoring rules; Appendix~\ref{app:systems} describes the compared systems.

\subsection[Output specification and settings]{Output specification and MLLM settings}
\label{app:eval-settings}

Table~\ref{tab:contract} defines the output space and scoring rule for the five downstream decisions, and Table~\ref{tab:mllm-settings} summarizes the evaluated MLLM settings. The default configuration uses GPT-5.6-Luna at low reasoning effort. All prompts are fixed zero-shot templates.

\subsection{Shared-perception protocol}
\label{app:eval-twostage}

The evaluation holds a shared perception pipeline fixed across systems. For each setting, the MLLM processes each episode in temporal order and produces a per-frame perception record. Every MLLM-based comparison system then receives this shared record and constructs its own cross-frame state and downstream predictions. Controlled variants apply the common decoder to \method{}'s fixed records and utterances with only the specified factor changed. All systems use the same output schema and scorer.

\subsection{Per-frame evidence and diagnostics}
\label{app:eval-store}

For each frame, the shared perception record contains the episode position; the set of active devices; the attended device, if any; a binary indicator of whether the user is attending to a device; the action, stated value, and queried topic expressed in the utterance, when applicable; and the requested action, together with a binary indicator of whether a request is present. Each system additionally receives the public device card, device roster, and current utterance.

For rendered episodes, prior activity observations determine when the perception pipeline extracts pointing cues, while the shared record provides the current-frame cue rather than accumulated engagement. Stated values come from the current utterance, and recall is produced during downstream prediction. Shared frames therefore yield identical inputs across the pair. Table~\ref{tab:perception} reports current-frame perception diagnostics for these shared records across the four MLLM settings.

\begin{table}[t]
\centering
\small
\caption{\textbf{Shared-perception diagnostics.} Current-frame active-screen set-$F_1$ and pointed-device accuracy on the rendered benchmark for the four evaluated MLLM settings (Table~\ref{tab:mllm-settings}).}
\label{tab:perception}
\setlength{\tabcolsep}{3pt}
\begin{tabular}{@{}l c c@{}}
\toprule
MLLM setting & \shortstack{Active-screen\\set-$F_1$ ($\uparrow$)} & \shortstack{Pointed-device\\accuracy ($\uparrow$)} \\
\midrule
Gemini 3.1 Flash-Lite (low) & 0.995 & 0.704 \\
GPT-5.6-Luna (low) & 0.992 & 0.882 \\
GPT-5.6-Luna (medium) & 0.991 & 0.900 \\
GPT-5.6-Sol (low) & 0.993 & 0.895 \\
\bottomrule
\end{tabular}

\par\vspace{2pt}\raggedright\footnotesize Active-screen set-$F_1$ uses all 2,400 frame instances; pointed-device accuracy uses the 1,600 pointing-frame instances.
\end{table}

\subsection{Scoring}
\label{app:eval-scoring-rules}

\paragraph{Deterministic decisions.} The construction fixes Eng, Inf, Rsp, and Nxt without model annotation; render metadata is used only for validation. Eng and Nxt use exact match, Inf uses case-insensitive exact match, and set-valued Rsp uses set-$F_1$, with the empty target matched exactly by the empty prediction. An Inf prediction of ``none'' receives credit only when ``none'' is the target.

\paragraph{Intent and overall performance.} Int is evaluated under one rubric by a fixed same-meaning judge, a text-only GPT-5.6-Sol at medium reasoning effort; the same judge scores every system, every MLLM setting, and the real-photo case study. The rubric credits paraphrases that preserve the action, target, device category, and material constraints. Empty or malformed predictions score zero. Overall is the macro-average of the downstream decisions.

\section{Comparison Systems and Variants}
\label{app:systems}

Alongside the compared systems in Table~\ref{tab:headline}, we evaluate controlled variants.

\subsection{Comparison protocol}
\label{app:systems-contract}

All systems follow the protocol in Appendix~\ref{app:eval-twostage}. Compared systems begin from the shared perception record, restricted to the input fields their designs specify, and retain their method-specific state-management mechanisms. Controlled variants share \method{}'s records, utterances, and decoder and change only the specified factor in carried state and, when applicable, current-frame input.

\subsection{Controls and baseline adaptations}
\label{app:systems-controls}

\paragraph{State controls.} \emph{Full context} retains the complete record and utterance history. \emph{Self-notes} updates a free-text memo at each frame, limited to 600 characters. \emph{Observation only} uses the current record. \emph{Answer cache} carries forward only the most recent answers, without counts. \emph{Text only} reads the utterances with no visual input; its near-floor engagement and recall confirm that the pair-separating evidence is visual rather than textual.

\paragraph{Baseline adaptations.} We adapt the memory and dialogue-state methods Mem0~\cite{chhikara2025mem0} and MM-DST~\cite{vdtn2022,olvit2024}, together with the multi-agent methods Debate-or-vote~\cite{choi2025debate} and Mixture-of-Agents~\cite{wang2025moa}. Mem0 maintains a fact memory built by extraction and consolidation and retrieved by lexical matching with a recency fallback; MM-DST maintains a slot--value state that keeps the most recent value and is read out each turn. Debate-or-vote coordinates per-device agents through peer revision and field-wise voting; Mixture-of-Agents synthesizes per-device proposals from per-proposer running summaries.

\subsection{Controlled variants}
\label{app:systems-ablations}

Controlled variants change only the specified state, evidence, or routing factor.

\paragraph{Variant definitions.} \emph{Frame-local state} resets the carried state before each frame. \emph{Count-free evidence} retains activity and pointing presence but removes their multiplicities. \emph{No pointing channel} removes pointing from carried and current-frame inputs while retaining activity evidence. \emph{Reduced-ability routing} removes only per-device ability clauses while retaining \method{}'s carried state and general card context.

Across the targeted contrasts, count and pointing evidence support Eng, prior state supports Inf, and per-device abilities support Rsp.

\section{Statistical Analysis}
\label{app:stats}

We evaluate overall-performance differences using a paired analysis clustered by matched pair. Table~\ref{tab:headline} carries the resulting gap intervals, and every adjusted bootstrap tail-area $p$-value falls below $10^{-3}$.

\paragraph{Endpoint and clustering.}
\label{app:stats-estimand}
For each sample, scored instances are first averaged within each downstream decision; Overall is then the equal-weight mean of Eng, Int, Inf, Rsp, and Nxt. Each contrast is \method{} minus the comparison, aligned by matched-pair identity and episode role. Resampling is clustered by matched pair, with the paired episodes retained jointly because the role exchange makes their outcomes dependent.

\paragraph{Designated frames.} For every system, Eng averages the shared request and recall frames, $f_{10}$--$f_{12}$, and Rsp averages the construction-designated routing frames, $f_9$ and $f_{12}$. Each resulting score enters Overall with the same weight as the other downstream decisions.

\paragraph{Bootstrap procedure and decision criterion.}
\label{app:stats-bootstrap}
Each of the $B=10{,}000$ replicates resamples matched pairs with replacement, retains the paired episodes jointly, and recomputes the Overall difference. If $k$ of the $B$ replicates yield a difference at or below zero, the one-sided tail-area $p$-value is the add-one estimate $(1+k)/(B+1)$. Conclusions are unchanged when resampling instead clusters the 46 rendered scenes that host the 100 pairs. The Gap brackets in Table~\ref{tab:headline} report pointwise percentile intervals from this distribution; the Overall brackets instead use a one-sample clustered bootstrap for each system. The paired-gap and one-sample interval procedures use the same fixed random seed. Holm correction is applied to the tail-area $p$-values.

The four adapted published-method comparisons under GPT-5.6-Luna at low reasoning effort are adjusted together with Holm's procedure. Full context, Self-notes, Answer cache, Text only, and Observation only form five additional comparisons that are adjusted separately. The decision criterion for the published-method comparisons is a Holm-adjusted $p<0.05$ together with an estimated difference of at least $0.10$.

\paragraph{Ablation contrasts.} The endpoint-aligned ablation contrasts are Holm-adjusted as a separate group; every adjusted $p$ falls below $10^{-3}$.

\section{Real-Photo Case Study}
\label{app:realcase}

\paragraph{Protocol and image preparation.} We construct one controlled matched pair from prepared photographs of a home-desk interaction involving a laptop, tablet, and phone; the user's face is not visible. All evaluated systems receive the same utterances, device card, roster, and closed answer vocabulary, and every system with visual input receives the same prepared photographs, while the episode script supplies the evaluation targets. Fixed display masks standardize screen state while preserving foreground hand and finger regions. Each shared frame is prepared once and reused across the pair.

\paragraph{Matched-pair construction.} The user compares restaurants on the laptop and tablet and states a reservation time for each. Subsequent hand-on-device contact favors the laptop in episode $A$ and the tablet in episode $B$. In the shared final frame, both work screens are dark, and the user asks from the phone, without naming a device, to text everyone the time for ``the one I was just working on'' (Figure~\ref{fig:app-realcase}). The shared setup, fact-stating, and request frames hold context fixed, while the intervening frames vary engagement and produce different engaged-device and recalled-time targets.

\paragraph{Results.} \method{} folds hand-on-device evidence from the prepared photographs into its carried state and predicts from that state and the current observation. Across the pair, it identifies the device with the largest accumulated engagement tally and recalls the associated time. Only \method{} also recalls the stated time on both episodes; the other designs, including Full context, do not (Table~\ref{tab:realcase}).

\end{document}